# Registering large volume serial-section electron microscopy image sets for neural circuit reconstruction using FFT signal whitening

Arthur W. Wetzel, Jennifer Bakal,
and Markus Dittrich
Pittsburgh Supercomputing Center
Carnegie Mellon University
Pittsburgh, Pennsylvania, USA
awetzel@psc.edu

David G. C. Hildebrand, Josh L. Morgan,
and Jeff W. Lichtman
Department of Molecular and Cell Biology &
Center for Brain Science
Harvard University
Cambridge, Massachusetts, USA

*Abstract*—The detailed reconstruction of neural anatomy for connectomics studies requires a combination of resolution and large three-dimensional data capture provided by serial section electron microscopy (ssEM). The convergence of high throughput ssEM imaging and improved tissue preparation methods now allows ssEM capture of complete specimen volumes up to cubic millimeter scale. The resulting multi-terabyte image sets span thousands of serial sections and must be precisely registered into coherent volumetric forms in which neural circuits can be traced and segmented. This paper introduces a Signal Whitening Fourier Transform Image Registration approach (SWiFT-IR) under development at the Pittsburgh Supercomputing Center and its use to align mouse and zebrafish brain datasets acquired using the wafer mapper ssEM imaging technology recently developed at Harvard University. Unlike other methods now used for ssEM registration, SWiFT-IR modifies its spatial frequency response during image matching to maximize a signal-to-noise measure used as its primary indicator of alignment quality. This alignment signal is more robust to rapid variations in biological content and unavoidable data distortions than either phase-only or standard Pearson correlation, thus allowing more precise alignment and statistical confidence. These improvements in turn enable an iterative registration procedure based on projections through multiple sections rather than more typical adjacent-pair matching methods. This projection approach, when coupled with known anatomical constraints and iteratively applied in a multi-resolution pyramid fashion, drives the alignment into a smooth form that properly represents complex and widely varying anatomical content such as the full cross-section zebrafish data.

## I. Introduction

To attain comprehensive understanding of information processing in the brain, we need to understand how neurons are interconnected and how their connectivity generates neuronal function. Structural data that are currently used to develop this understanding are captured by a variety of imaging techniques over a range of resolutions and from both fixed and *in vivo* specimens. An overview of imaging methods and challenges involved is provided by Lichtman and Denk [1].

Reconstructive mapping of detailed wiring in local brain circuits, or even the entire brain of small animals currently, at EM resolution requires massive volumes of image data. The need for such large data sets comes from the combination of long branching neural pathways, typically spanning hundreds of microns, together with the high resolution needed to reliably map axons and dendrites down to ~50 nm diameter. The summary from a 2011 neuroscience mini-symposium describes several state-of-the-art projects using large-scale automated histology methods to study links between brain structure and function [2]. These papers further explain the need for high-resolution neural circuit reconstruction using ssEM imaging.

An example of one section in a 10,000 section 100 Tvoxel ssEM dataset from mouse visual thalamus, described in [3], is shown in Fig. 1 below. The full section, shown at upper left in ~300x reduction, is stitched from 16 raw image tiles. The full area of each section contains near molecular level detail as apparent by the nuclear membrane near the top of the lower right image that is composed of two lipid bilayers. The entire physical section area, ~400 microns square, was imaged at 4nm/pixel sampling to produce this 10 Gpixel plane along with 10,000 other similar sections aligned into the entire 3D dataset.

Registration is an essential first step to transform stacks of raw ssEM images, such as Fig. 1, into coherent volumetric forms prior to 3D content analysis. High quality alignment is critical for the speed and accuracy of the later content analyses that have to follow rapidly changing biological structures within the assembled 3D volumes. Ideally we would align an entire dataset before content analysis begins; in practice, though, there is always some preliminary analysis even while an image set is being acquired in its entirety. Results from these analyses identify damaged sections, areas of systematic distortion and provide useful constraints that help to guide and accelerate registration processing.

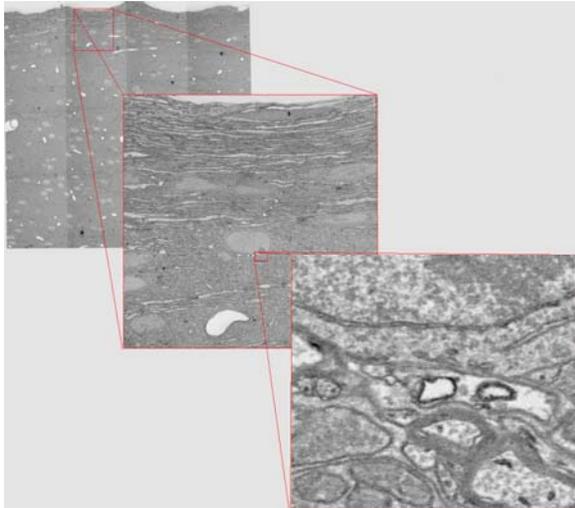

Fig. 1 Progressively zoomed views of a single 10 Gpixel section within a 10,000 section 100 Tvoxel ssEM stack from mouse visual thalamus. The images are progressively 400 microns wide at top left, 80 microns middle and 2.5 microns at bottom right.

## II. Background

There are numerous approaches for registering images according to various content and purposes, of which this paper mentions but a few. Many registration codes derive from a least squares intensity differencing approach first proposed 35 years ago by Lucas and Kanade at CMU [4] and later refined into the Kanade-Lucas-Tomasi (KLT) feature tracker [5]. These methods can be very effective when changes of content between adjacent pairs are small and intensities well matched; hence, they are widely used in 2D tile stitching and video stabilization. In our experience, also noted by others working in this area [6], they are less effective for ssEM registration due to multiple optima in the difference function and excessive sensitivity to rapid changes of adjacent image content, artifacts, and overall variations of intensity and contrast.

Most other registration codes use either normalized cross-correlation (NCC), also called Pearson correlation, or phase-only correlation (POC). The computations are usually done in the frequency domain to obtain the speed advantages provided by the FFT. Although FFT correlation, performed by inverting the frequency domain multiplication of the two input images, does not directly provide the ±1 range of the Pearson correlation coefficient, a simple cumulative area correction produces the equivalent result when required [7]. Many researchers begin with the assumption that NCC is the gold standard metric of image matching. General familiarity with statistical correlation makes it the most widely chosen method for image matching although, as discussed shortly, it has severe limitations in practice.

Frequently, scale invariant feature transforms (SIFT) [8] or speeded-up robust features (SURF) [9] are used to locate initial corresponding key-points between pairs of images. Along with a search procedure, such as RANSAC [10], these points help to initialize the larger scale match process by producing an initial approximate alignment. To gain efficiency these methods are usually implemented using a multi-resolution hierarchy in which the coarsest levels are solved first and successively feed into higher resolution solutions. In the computing field this is called a pyramid or, more recently, a multi-grid strategy.

The principles of the above methods, including the KLT and FFT-based registration, have been incorporated into many programs. The National Library of Medicine Insight Segmentation and Registration Toolkit (ITK) [11][12], primarily targeted to biomedical applications, provides open source versions of a variety of registration and segmentation algorithms. Elastix [13], for example, is a deformable registration toolkit that is very useful for aligning stacks of single images, most commonly CT and MRI, that is implemented using ITK.

Several registration approaches have been taken by researchers at the HMMI Janelia Farm research campus. TrakEM2, http://imagej.net/TrakEM2, originally developed by Albert Cardona while at the Institute of Neuroinformatics in Zurich and now maintained by Cardona's group at Janelia incorporates viewing, user driven circuit tracing, and Pearson-correlation-based registration. Stephan Saalfeld and Cardona have also worked on several alternative techniques [14][15]. Louis Scheffler, Bill Karsh and others have developed a large scale parallel processing pipeline used for EM alignment of drosophila datasets [16].

Tolga Tasdizen and colleagues at the Scientific Computing and Imaging Institute at the University of Utah have developed their NCRToolset [6] which has been used for volumes up to 15 Tvoxels. Its base level FFT correlation uses an image filtering process to reduce the number of local minima. The Utah software achieves a differential frequency effect as a result of median and Gaussian smoothing which helps to improve convergence. However, in some cases that smoothing may limit the final registration precision which ultimately depends on high frequency Fourier coefficients.

Work by our PSC group in 2009 aligned a 4nm resolution 10 Tvoxel mouse visual cortex volume [17] from transmission electron microscopy data and a recent 100 Tvoxel volume using the same capture and registration methods [18]. This alignment software, AlignTK written by Greg Hood, can be downloaded from http://www.mmbios.org. AlignTK is based on Pearson correlation and stochastic hill-climbing, combined with a spring mesh relaxation technique for global smoothing.

Despite this range of alignment tools, unresolved issues remain due to critical parameters such as spring mesh tensioning, multimodal image match signals with Pearson correlation, lack of robustness with phase only correlation, and relatively slow computational speeds when stochastic methods such as RANSAC are used or when iterative methods have slow convergence. All of these issues become more severe as data size grows and particularly when faced with extreme variations of content and the need for geometrical fidelity as exemplified by the zebrafish alignment described later in this paper. Therefore, it is useful to reexamine the core mechanisms that can be used for computational registration and to test their utility with current leading edge ssEM datasets.

## III. METHODS

The SWiFT-IR method described in this paper is motivated by several factors in addition to those just mentioned. First, it is well known from communications and RADAR theory that the maximum likelihood detection and time localization of a specific signal template, such as a satellite GPS signal, in the presence of additive white gaussian noise is given by a matched filter [19]. That is, the correlation of an unknown signal with the shape of the template as it would be distorted by both the systematic characteristics of the communications channel and the additive noise. This principle is built into many systems when low-level, known signals must be robustly detected in the presence of noise. The proper correlation can be done in the frequency domain by manipulating the frequency responses to equalize the noise contributions across all frequencies[20]. Although most often stated as a time domain problem, this principle also applies to higher dimensional signals such as our ssEM 2D regional template matching [21].

The outputs of matched filters are evaluated by their peak signal-to-noise ratio (SNR) relative to statistical background fluctuations. In the ideal case of an identical noiseless image pair containing no repeating content, the POC pushes the correlation energy into a single Dirac delta function [22]. When one of these identical images, hence noiseless with respect to each other, is shifted relative to the other, the Dirac correlation pulse appears in a position which indicates the relative image shift. Multiplying the Fourier transform of the shifted image by the repositioned Dirac pulse's Fourier transform will then shift the displaced image back into proper correspondence with the other. Images with repeating content will produce multiple peaks corresponding to the repeat positions. While this elegant behavior is not typically achieved in practice it is nevertheless important for understanding the underlying principles of image registration.

Real images include noise and distortions which reduce the ideal Dirac pulse into a broadened signal that must be detected and localized. When content variations and noise reach a sufficiently high level, the correlation peak or peaks become undetectable and fade into background noise. The ultimate limits of registration accuracy are given by the Cramer Rao bound (CRB) [23][24], which is a theoretical limit of minimum alignment variance that could be achieved by any procedure. The CRB is a function of image content and noise variance.

As described by Pham et. al.[25] for uncorrupted images with zero mean additive Gaussian noise the CRB can be realized by coarse-to-fine hierarchical registration. In the case of ssEM we clearly do not have uncorrupted images or additive white noise. From a mathematical perspective the generally large change of biological content from one section to another imposes the equivalent of a large non-Gaussian noise. Nevertheless, the Pham paper provides guidance by showing that quickly converging, hierarchically iterative alignment methods can be implemented in practice using image area based correlations. Area correlations avoid the need for preliminary feature identification, such as SIFT or SURF, and also avoid stochastic search procedures but their computations should be limited to areas of actual image content rather than regions of smooth background.

Importantly, CRB analysis does not include the effects of anatomical pose when structures may be oriented differently than the lowest energy image match may indicate. Any ssEM image set captures an instantaneous configuration of the anatomical content. At the smallest scales many features such as, in the case of connectomics image sets, synaptic vesicles and mitochondria would, *in vivo,* be in rapid motion and the overall bend of a flexible animal such as the zebrafish is different than may be indicated by their slice-to-slice correlations or than would be achieved by long range relaxation procedures. Therefore, we need external objective guiding information in order to produce anatomically correct registrations. That information may be limited to symmetry considerations or may take into account the known shape of the fixed specimen block prior to sectioning, shape information from low resolution prefixation optical images or, in the best case, micro CT or electron beam tomography of the specimen within the block. This type of information is particularly important if the purpose of study is to characterize distances within a specimen or phenotypic shape variations.

Another motivation for the SWiFT-IR approach is the excellent quality that had already been achieved many years ago in optical domain signal processing. Discussions with the late Dr. Emmett Leith in conjunction with another project included review of his early work in nonlinear joint transform optical correlation achieved by the combination of optical domain Fourier transforms together with nonlinear optical sensing [26]. Related color pattern correlation work by Nicolas and Campos is included in a tribute publication to Leith and Denisyuk for their contributions to holography [27].

The optical correlator that relates most directly to SWiFT-IR's process is by Bahram Javidi [28] which demonstrates the large gain in two-dimensional signal detection and localization, particularly for "colored" noise (non Gaussian) situations, that is achieved using an easily implemented exponential power scaling of Fourier transform components. This is extremely important for ssEM registration because we have highly non-Gaussian "noise" due to variations in biological structure from section to section, the similar appearances of different cells and other common biological structures, along with artifacts and unknown distortions of the tissue. These effects are much larger than image capture noise in the classical sense.

In the frequency domain, represented by the components of a Fourier transform, this whitening process accentuates higher spatial frequencies and therefore sharpens edges. Signal whitening in natural vision also performs edge enhancement and localized decorrelation of information in the visual system. Although Javidi's team implemented their pattern matching procedures directly in the optical domain [29][30], the same mathematics applies to computation using FFT correlation but with nonlinear digital manipulation of the magnitudes of the individual frequency components. We can make a simple analogy to an audio tone control to accentuate treble or bass or finer tonal control by a multi-channel audio graphic equalizer.

Although it is well known that the phases of Fourier image representations contain substantially more important visual content than the amplitudes, it is significant that nonlinear whitening, as just discussed, modulates only the amplitudes and does not modify the phases which contain the relative shift information that registration processes need to uncover. In fact, power based amplitude scaling does not directly use the frequency of the Fourier coefficients. Rather, it systematically affects signals according to frequency only because, like nearly all natural images, ssEM image content has much larger low frequency content which falls off at higher spatial frequencies.

Nevertheless, these power based amplitude changes have a large impact on the effectiveness of correlation processes which depend on the underlying phase shifts. SWiFT-IR takes advantage of the frequently overlooked importance of using a broad range of whitening levels from normalized correlation at one extreme to POC at the other. NCC is a special case with no signal whitening in which original amplitudes are preserved, while POC is the most extreme whitening which normalizes all frequencies to a uniform amplitude. SWiFT's implementation controls this behavior using a 0 to 1 scalar parameter as the exponent for an amplitude power function. In SWiFT-IR computations, this is adjusted to optimize the matching process giving a more robust foundation for automated registration.

Fig. 2 below illustrates the whitening parameter's effect when locating damaged region B in undamaged area A. The three bluish images show match responses for NCC at left, SWiFT center and POC right. White is a strong signal, black is negative correlation, and blue is no match. NCC, at left, does not properly locate the best match position and presents a dilemma of how to use its smeared result. Although hard to see in printed form SWiFT generates a single bright response indicating the proper match location with minimal response elsewhere while POC, shows a less distinct lower score at the proper position along with other scattered ambiguous speckles.

The excessive emphasis of high frequencies makes POC intolerant of geometric distortion. This is because distortions and biological variations, which are large relative to normal high frequency Fourier terms, push the high spatial frequencies out of phase.. If the relative shape of Fig 2 A and B had been slightly worse the match position would be lost in background speckles. Therefore POC is not very useful for section-to-section image matching due to its unacceptable failure rate.

As the SWiFT-IR name implies, the technique can be implemented to run quickly due to its low computational complexity and the availability of highly optimized FFT codes. We generally process large ssEM images as a tiled series of local patches rather than as full area images. This reduces the size of the individual FFTs, also improving speed. This computational framework is enhanced, relative to usual FFT correlation, by performing adjustments of the amplitudes of the transform components, either before or after the frequency domain multiplication that is the core of FFT correlation. Each of these whitening positions has its own operational advantage while still producing equivalent results.

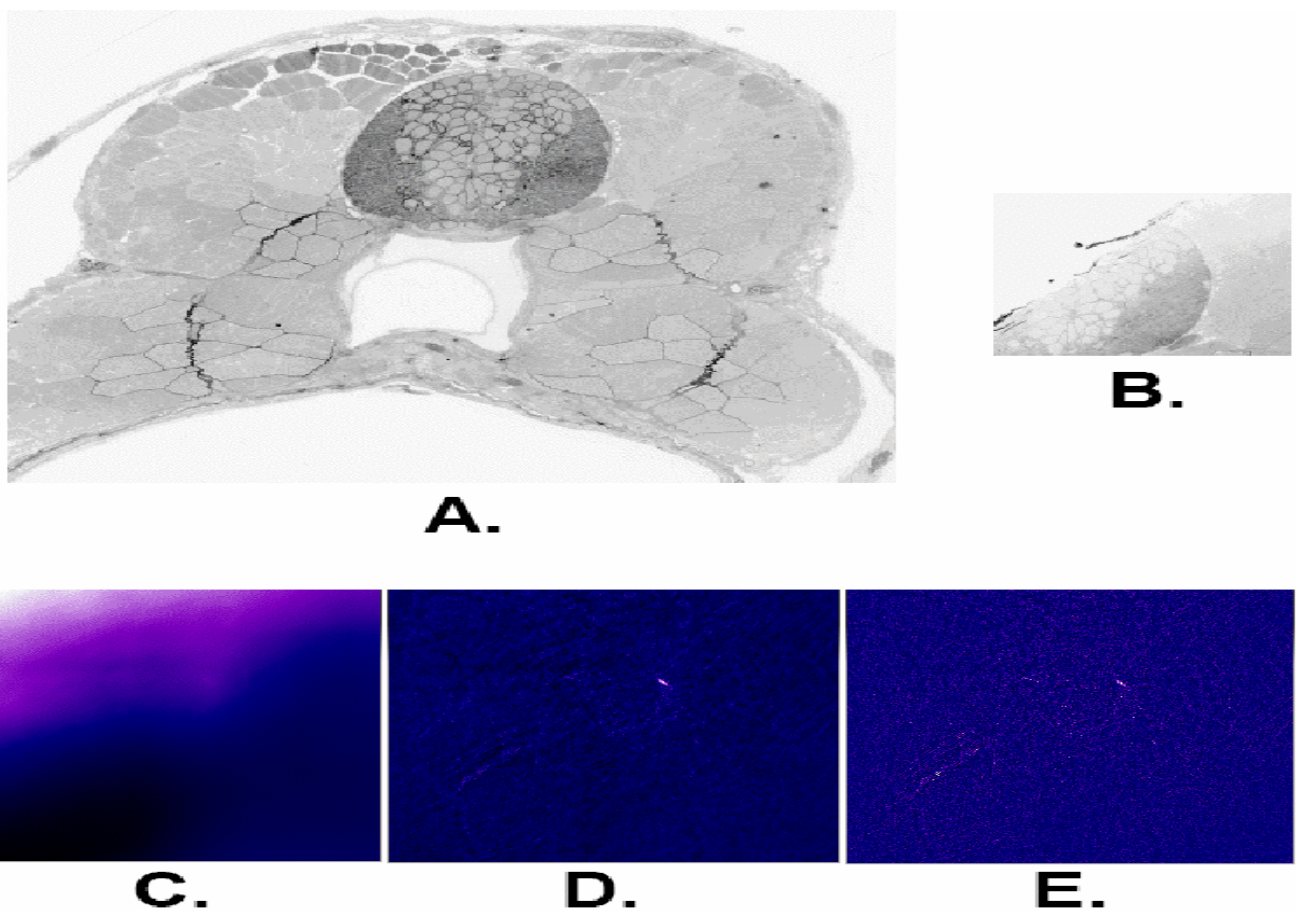

Fig 2. Correlation responses for whitening levels of 0, NCC, at left 0.7 middle and 1.0, POC, at right. See discussion in text.

Pixel level match responses, such as Fig 2, are easily converted to Z scores repreresenting the number of standard deviations (i.e. sigma) away from the background distribution. This SNR is the most important information available for making signal detection decisions relative to the background distribution. When a particular value has a sufficiently high SNR then it is statistically likely to be the proper match. In Fig 2 the SNRs are respectively 2.0, 13.5 and 8.5.

With extremely large ssEM image sets some number of individually unlikely events will occur by chance. SNR values help us to assess the likelihood that we have real matches rather than statistical outliers. To put this in perspective, a one tailed SNR of 6 sigma indicates, from the Q distribution of Gaussian tail probabilities, a 1 in $3 \times 10^9$ chance of being a statistical outlier. SNRs greater than 15, as frequently obtained even in the early stages of SWiFT-IR alignments, are excellent indications that we have correct matches and, unlike with NCC or POC, we can stop search processes to avoid time consuming and fruitless searches for even better matches. In later stages of registration we often see SNRs of 80 to 100. Such extreme levels indicate fat-tailed non-Gaussian distributions rather than random fluctuations but nevertheless indicate relative quality.

These previous concepts, matched filtering and power scaling of Fourier coefficients, correspond to key elements that distinguish the SWiFT-IR approach. Although we do not have the proper circumstances, known signal and additive Gaussian noise, to directly implement matched filtering we can make a step in that direction by noting that in the absence of gross defects the correlation within well-aligned regions of ssEM stacks between any particular section and the average of its neighboring sections in the Z direction perpendicular to the cut planes is stronger than to either of its immediate neighbors alone, whether done by NCC, POC or SWiFT's SNR measure.

In part this improvement is due to the effectiveness of noise reduction by the averaging process as $1/sqrt(N)$ where N is the number of averaged sections. Most of the image matching for SWiFT-IR uses this type of local average as a "model" template so that new parts of raw data are matched to the current model rather than to parts of adjacent raw data. In practice it is important to omit the section corresponding to the current raw data from its surrounding model and to also avoid putting severely damaged sections into the model. The number of sections N used in the model is typically reduced as registration progresses.

The averaging of aligned sections provides a projection in Z that preferentially emphasizes features that are relatively perpendicular to the cutting plane and also provides natural continuity along linear and gently curved but obliquely oriented structures as they cross the corresponding raw sectioning planes. During early stages of registration, which are generally done at reduced resolution, a large number of sections can be used in the model to automatically get the desired global smoothing effect that would otherwise be done as post-processing steps using thin-plate splines [31], spring meshes, or other iterative relaxation mechanisms.

Another important aspect of the model method is that we obtain an implicit parallelism since a given region of raw data usually does not have to be individually matched to each of the sections in the current model's span. With rare exceptions the match process automatically incorporates the effects of the model's Z averaging range all at once. Exceptions are cases of abrupt content or magnification changes where the projective match can become locked to one side or the other resulting in breaks within otherwise smooth alignments. These jumps can be fixed using a slightly more expensive matching procedure whereby any offending sections are matched separately to the models on either side of a jump so that a Bezier or other smoothing spline can be built across the failed region.

In practice, many registration procedures have substantial complexity and computational overhead to compensate for the limited robustness of NCC and POC. As previously noted, the Utah registration process uses a filtering technique that helps reduce false matches. In our experience many available codes do not take advantage of a particularly useful form of spatial domain filtering that, in optics, is known as apodization. Apodization tapers the contrast of match region edges to avoid false matching effects that are particularly notable with FFT processing due to wrap around in the frequency domain. The SWiFT-IR technique applies a cosine taper to prevent this problem. This is slightly different from a windowing function, such as the often used Blackman window, since the cosine taper preserves more of the image content that is inside the local image blocks and therefore maintains higher sensitivity.

The ability of SWiFT-IR signal matching to properly to identify and localize corresponding image regions for virtually all areas of usable image content simplifies our registration process so we need only a few relatively simple program modules for image conversion, scaling, matching, rendering and averaging. We currently implement these by programs respectively named **icon**, **iscale**, **swim**, **mir** and **remod**.

**Icon** converts from initial 16-bit to 8-bit forms and, during that process, also adjusts intensity and contrast. **Iscale** produces the hiearchical source pyramid that is used during alignment processing. This is usually by powers of 2 although multiples of 3 and 5 are also useful. **Swim** implements the signal whitened image matching process. **Mir**, multiple image rendering, produces aligned output images according to linear least squares solution of correspondence points found by **swim**. This is implemented as a software texture mapping process applied to a triangulated mesh. Each triangle is locally affine but higher order curved representation, Bezier or other splines, can be generated by triangle subdivision. Finally, **remod** generates locally Z averaged models which, as noted earlier, provides geometric smoothing along Z and noise reduction.

Registration begins with highly reduced images, the base of the pyramid hierarchy, selected to be small enough to allow very fast processing but large enough so that large structures, which are implicitly the alignment features, can be seen by both automated and visual means and in which the overall alignment pose can be established. In ssEM data, large structures include capillaries, blood cells and even cell nuclei roughly 10 microns across. When viewed at one micron per pixel sampling, these features are easily visible, so this is a convenient starting scale. The exact value is not critical. With our data, one micron per pixel corresponds to reduced scale full section images a few hundred pixels wide.

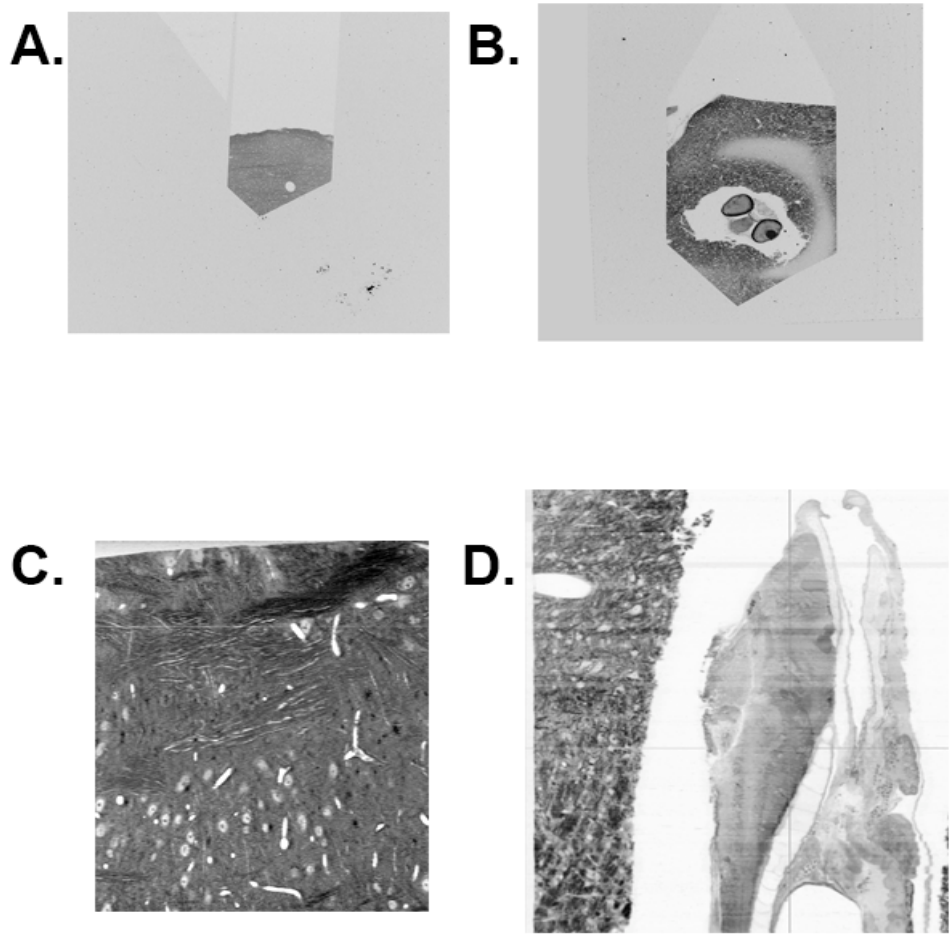

Fig 3. Examples of low resolution full block face images at top and corresponding preliminary 3D reconstructions.

With wafer mapper datasets we have an additional source of data that is very useful for building the starting pyramid. These data are complete sets of low resolution overview images. In their original purpose these images, such as the A and B examples shown at the top of Fig 3, were used for locating sections on the wafers and determining coordinates for controlling the stage position and rotation during later high resolution scans. Typically the high resolution scans are limited to user-selected regions of interest (ROI) that are much smaller than the full block face to reduce the already large volume of image data and speed up the high resolution scanning process.

Due to their low resolution and generally lower quality, it was originally expected that there was no further use for the overview images. However, there are several valuable aspects of the overviews that are not present in the high resolution scans. First, as mentioned, the overviews capture the entire block face. This is important because the block edges are the most objective indicators of specimen shape when blocks are trimmed into known size and shape. This information transfers directly into making high quality approximations of interior ROI configurations. Second, each section overview is captured in a single untiled image so there is no issue of ROI drift whereas the limited precision of ROI targeting always leaves ragged edges around the full resolution data. Third, despite higher per pixel noise the full area intensities of overview images are more consistent than corresponding high resolution scans and particularly for tiled datasets. This overview information can be easily fed forward to constrain intensity and contrast equalizations on the full resolution data. Finally, since the full overview set is captured before high resolution scans are started, an initial alignment of the overview is useful for establishing the specimen pose, for identifying regions with artifacts or other problems, and for accurately selecting the regions of interest needed for later full resolution analyses.

Images C and D shown above in Fig 3 are perpendicular cuts through the initial SWiFT aligned volumes of the low resolution overview image stacks for LGN, left image C, and zebrafish, image D on the right. Notice that in both cases, despite being viewed edge on to the cutting planes, there is coherent and realistic image detail. The largest white features in the LGN are capillaries and the light oval shapes are neuron cell bodies. The vertical scale is very compressed and spans the entire 10,000 section stack. Similarly in the right side image D the preliminary low resolution zebrafish alignment through 16,000 sections properly displays the outer structure of the fish along with its appropriate internal structure. The dark area on the left is external supporting material, mouse brain in this case. Most critically, the left edge that fades into white is the straight edge of the tissue block as trimmed prior to sectioning. That edge is our most objective external reference governing vertical shape of the reconstructed fish body. Faint horizontal streaks across the fish are variations of section intensities and contrasts that were left uncorrected during this early alignment.

## IV. RESULTS

The result of a large scale SWiFT-IR registration is best summarized by Fig 4. The large left side image is a coronal cut through the length of an aligned 16,000 section 60nm/pixel zebrafish dataset. Each of the 16,000 dewarped and **mir** rendered section images that are stacked on edge in this view are 9,500 by 8,600 pixels. Therefore, this particular cut is just one of the 8,600 possible coronal cuts. The image is reduced along its Z length from 16,000 to 2,000 pixels for printing and the horizontal direction proportionally scaled by 8x. Clearly there is substantial water filled, hence blank, space within the fish and outside of the body. Therefore the bounding box volume is several times larger than the number of fish voxels.

The most important point of Fig 4 is that the overall shape of the fish including eyes, brain structure and even the fins, at the bottom left and right, are well aligned. Similar quality extends over the entire volume. The expanded regions on the right are full resolution zooms showing that registration fidelity is very good even at maximum detail.

The full zebrafish dataset consists of four scans. The first, as discussed in relation to Fig 3, was the full low resolution overview series of 18,200 sections. Preliminary wafer mapper analysis of the overview resulted in microscope positioning coordinates for higher resolution scans. Based on biological content and some damage at the beginning of the sectioning process a primary data set of 16,000 sections was reimaged at 60nm resolution covering full body cross sections plus a 12,400 section span encompassing the entire brain area was also imaged at 20nm per pixel. At 20 nm per pixel there was a 3:1 aspect ratio between the fixed 60nm section thickness and the pixel spacing. An 800 section span targeting one pair of neuromast sensory organs, was acquired at 4nm per pixel.

Zebrafish registration started with the overview image set. Because overview images cover the full block face the shape of the block edges provided a convenient approximate alignment that was refined by several iterations of the SWiFT software. These initial alignments used the full block area, including mouse brain support tissue, and did not separately look at the zebrafish content.

After this preliminary step it was now possible to determine the relatively consistent systematic distortion of the fish blody due to compression during sectioning. Because the fish happened to be positioned diagonal to the cutting direction cutting compression introduced an obvious shearing of the fish anatomy. By measureing the angle between the vertical symmetry axis of the fish in a few areas, such as the eyes, that are known to be horizontally level in normal zebrafish anatomy the first estimate of shear could be computed and then applied to renormalize the shape of the edge based alignment volume.

The next refinement pass used this geometrically corrected volume as a model for rematching onto the full raw overview dataset. Thus we were already molding the reconstruction shape into a canonical pose that would not have been produced by typical registration processing. At this low resolution stage there was no need for triangulation. Full section area affine registration was sufficient. This process of image matching, rendering, and remodeling was iterated until there was no significant change in SNR scores from the **swim** process.

At this initial stage, while the data was small and while debugging the process, it was very useful to review the entire registration output. The **qiv** program that is normally available with Linux allowed fast section-by-section viewing at up to 25 frames per second. At that rate it was easy to review the entire 16,000 image stack in about 10 minutes. A small modification to **qiv** recorded cursor and click positions. This was useful for quickly marking damaged and improperly aligned sections during high speed viewing. Additional visualization of aligned data as arbitrarily oriented 3D cut-planes used the PSC Volume Browser (**PSC-VB**) that we initially developed for use for the National Library of Medicine Visible Human and other large volumetric datasets. User controlled 3D navigation in **PSC-VB** allowed targted visual examination of volumes and regional areas to evaluate the correctness of anatomical structures.

The main SWiFT-IR workflow is to iteratively align at some current working resolution until the SNR scores stabilize and then advance to the next higher resolution by first expanding the current model and then matching new higher resolution to that model. The most computationally intensive part of this process, **swim**, runs at ~10 Mpixels/sec per core using single precision FFTW3 on our 3.3 GHz 32 core Intel E5-4627 machine with 512GB of shared memory.

Typical undamaged zebrafish sections are sufficiently well formed that we were able to use affine registration all the way to the full 60nm resolution. At this final stage many residual nonlinear deformations, that were not accurately determined from lower resolutions, had to be corrected. This was done using a locally affine triangulation mesh to approximate the overall nonlinear warping. In this process, each affine triangle is produced from the linear least squares solution of swim match points within that triangle's area and also within an external band that includes the six surrounding triangles. Although most sections approached their final form with only one step of triangulation there are damaged or distorted areas that required more stages of nonlinear processing. In particular several spans of alternating thin and thick sections required four triangulation passes at 60nm plus three more to stabilize the 20nm data into consistency with the 60nm data.

At the end of the process we have all four image resolutions, overview, 60nm/pixel, 20nm/pixel and the small 4nm/pixel dataset, all aligned to the same global coordinate system. There were approximately 300 sections out of the 16,000 that have more severe problems such as cuts, tears and other defects that required additional manual corrections.

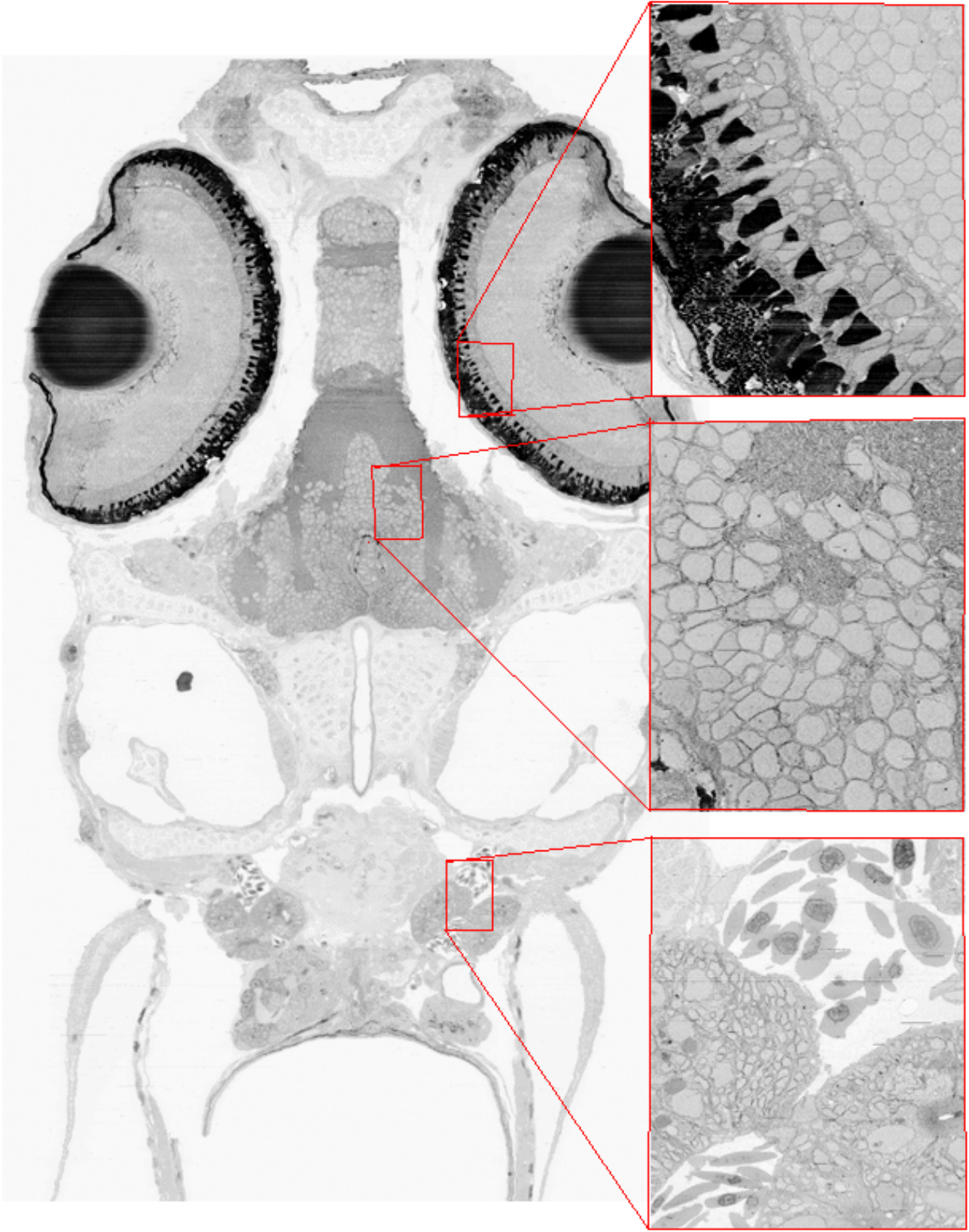

Fig. 4. Greatly reduced coronal view of the 16,000 section 60nm/pixel zebrafish data with three representative full resolution cut outs on the right.

## V. CONCLUSION

In summary, our experiences during development of the SWiFT-IR approach have shown its effectiveness with two large wafer mapper datasets. We have demonstrated that the robustness of the tunable signal whitening technique provides substantial advantages, particularly in regions of damaged or otherwise marginal image content, and simplifies key parts of the registration process. Of course the method can not reproduce missing data and there were cases, most notably encountered with the LGN dataset, where the density of similarly oriented cracks in the pick up tape were not properly handled by the SWiFT-IR code at that time. In the later zebrafish data there were initially difficulties in the vicinity of extremely high contrast features in the lens and retina where tissue staining is highly variable and around the outer surface of the fish. These issues were largely part of the learning curve of discovering how to best apply the elements of the SWiFT method and particularly the importance of rejecting damaged and highly distorted regions during model building operations.

The great benefit of wafer mapper overview images was a pleasant surprise. Its clear that, despite their limited quality and resolution, the availability of full block face imagery to establish objective references useful for correctly shaping the global alignment is superior to any result that would come from iterative relaxation methods. Particularly for the zebrafish this reference to overview data was the only way to be certain that bumps and curves in the reconstruction are real rather than registration artifacts. This gives added confidence that we have high anatomical fidelity throughout the entire specimen.

It is also clear that providing improved low resolutions of the quality we achieved would be of great value in improving ROI targeting consistency for high resolution scans. Achieving high quality alignments from low resolution scans is further enhanced by having textured background across the full block face surrounding an embedded specimen. In the zebrafish data this was approximated by the supporting mouse brain tissue. We also suggest that with enhanced stage control and modulation of the SEM scan raster the direct acquisition of high resolution data in high quality affine alignment would be possible either as tiles or, by using dynamic stage motion, as long pushbroom column scans. This would greatly reduce the issues of ragged edges in the 3D reconstruction volumes.

Finally, we have also seen there is still an important role for a human in the loop process. For the zebrafish this was most apparent in establishing a consistent bilateral centerline that would put the fish into a canonical straight line pose. An initial estimate from SWiFT matching of left right image flips was useful in the early stages. However, at higher resolution there were regions of asymmetry in this particular specimen where manual guidance gave a better result. Manual observation and intervention was also important in recognizing unexpected effects of alternate thin and thick section cutting that produced alternate directional bowing along the cut direction. In this case the automated model process produced jumps rather than a smooth result. This was solved by manually removing the thin section images from the model. It would be straightforward to automate this process in the future if it becomes a common problem with other datasets.

Further work is underway to make SWiFT-IR codes more user accessible through an appropriate user interface and also to greatly enhance the speed through both algorithmic means, such as pruned FFTs, and technical means including large memory parallel computers and OpenACC GPU acceleration..

ACKNOWLEDGMENT

Research reported in this publication was supported by the NIGMS of the NIH award number P41GM103712, D.G.C.H. NIH training fellowships (4T32MH20017, 4T32HL007901) and J.W.L. NIH (1U01NS090449, TR01 1R01NS076467)